\documentclass[10pt, journal, twoside]{IEEEtran}
\usepackage{amsmath}
\usepackage{graphicx}

\usepackage{multirow}

\usepackage[colorlinks,linkcolor=blue]{hyperref}
\usepackage{tabu}

\usepackage{subfigure}
\usepackage{amsfonts}
\usepackage{soul}
\soulregister\cite7 % 针对\cite命令
\soulregister\cref7 % 针对\cref命令
\soulregister\footnote7 % 针对\footnote命令
\usepackage{multirow}
\usepackage[table,xcdraw]{xcolor}
\usepackage{color}
\usepackage{amsmath}
\usepackage{amssymb}
\usepackage{mwe}
\usepackage{paralist}

\usepackage{amssymb,cool}
\usepackage{multirow}
\usepackage{booktabs} %table
\usepackage{stfloats}
\usepackage{pifont}
\usepackage[normalem]{ulem}
\usepackage{cite}

\usepackage{algorithm}
\usepackage{algorithmic}
\usepackage{color}
\usepackage{url}

\usepackage{url}

\usepackage[capitalise]{cleveref}
\usepackage{color}
\usepackage[vlined,linesnumbered,ruled,algo2e]{algorithm2e}
\usepackage[linesnumbered,ruled,vlined]{algorithm2e}
\usepackage{booktabs}
\usepackage[table,xcdraw]{xcolor}

\begin{document}

\title{GaitMPL: Gait Recognition with Memory-Augmented Progressive Learning}

\author{Huanzhang Dou$^*$, Pengyi Zhang$^*$, Yuhan Zhao,\\
 Lin Dong, Zequn Qin, Xi Li
\thanks{*Equal contribution.}
\thanks{Xi Li and Zequn Qin are co-corresponding authors.}
\thanks{Xi Li is with Zhejiang University, Shanghai Institute for Advanced Study of Zhejiang University, and Shanghai AI Lab.(email: xilizju@zju.edu.cn)}
\thanks{H. Dou, P. Zhang, and Z. Qin are with College of Computer Science, Zhejiang University, Hangzhou, China (email: \{hzdou, pyzhang, zequnqin\} @zju.edu.cn)}
\thanks{Y. Zhao is with School of Software Technology, Zhejiang University, Hangzhou, China (email: yuhanzhao@zju.edu.cn) }
\thanks{Lin Dong is with Merit Interactive Co., Ltd., Hangzhou, China (email:dongl@getui.com) }
\\}

\maketitle

\begin{abstract}

Gait recognition aims at identifying the pedestrians at a long distance by their biometric gait patterns.
It is inherently challenging due to the various covariates and the properties of silhouettes (textureless and colorless), which result in two kinds of pair-wise hard samples: the same pedestrian could have distinct silhouettes (intra-class diversity) and different pedestrians could have similar silhouettes (inter-class similarity). 
In this work, we propose to solve the hard sample issue with a Memory-augmented Progressive Learning network (GaitMPL), including Dynamic Reweighting Progressive Learning module (DRPL) and Global Structure-Aligned Memory bank (GSAM). Specifically, DRPL reduces the learning difficulty of hard samples by easy-to-hard progressive learning. GSAM further augments DRPL with a structure-aligned memory mechanism, which maintains and models the feature distribution of each ID. 
Experiments on two commonly used datasets, CASIA-B and OU-MVLP, demonstrate the effectiveness of GaitMPL.
On CASIA-B, we achieve the state-of-the-art performance, i.e., 88.0\% on the most challenging condition (Clothing) and 93.3\% on the average condition, which outperforms the other methods by at least \textbf{3.8\%} and \textbf{1.4\%}, respectively.
  
\end{abstract}


\section{Introduction}

\begin{figure}[t]
    \centering
    \subfigure[The two types of hard samples. The sequences of the first two rows show the intra-class \textbf{diversity} and the sequences of the last two rows show the inter-class \textbf{similarity}. The diversity and similarity of silhouettes incur \textbf{intra-class hard positives}  and \textbf{inter-class hard negatives}, correspondingly.]{
        \begin{minipage}[b]{0.4\textwidth}
            \centering
            \includegraphics[width=0.9\textwidth]{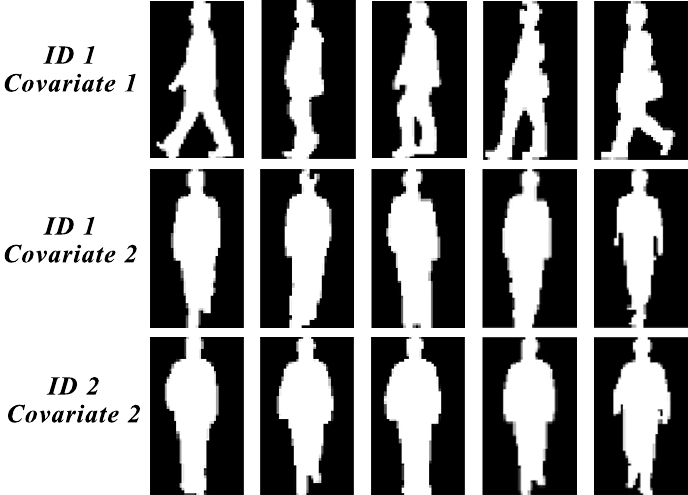}
        \end{minipage}
        \label{shoutu1}
    }
        \subfigure[The illustration of the different granularities of visual clues between easy/hard negatives. Easy negatives could be distinguished by evident visual clues while hard negatives have to be distinguished by subtle visual clues, e.g., the arms. To highlight the visual clues, they are marked with different colors.]{
            \begin{minipage}[b]{0.4\textwidth}
            \centering
            \includegraphics[width=0.95\textwidth]{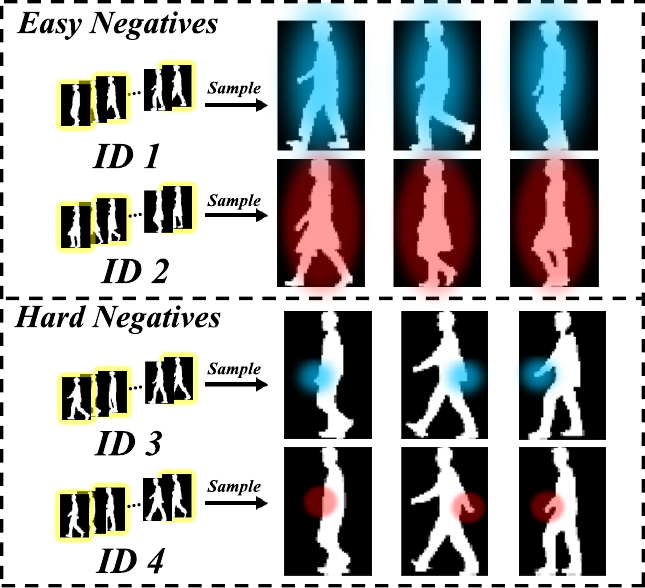}
            \end{minipage}
        \label{shoutu2}
        }
    \caption{The illustration of two types of hard samples and the different granularities of visual clues between easy/hard negatives.}
\end{figure}

Appearance-based gait recognition is one of the most promising biometrics, which represents the walking pattern of pedestrians and is extracted from silhouette sequences. Different from other biometrics, gait can be recognized at a long distance without the explicit cooperation of pedestrians. Thus, gait recognition has drawn increasing research attention and reveals great application potentials recently, such as security checks~\cite{chattopadhyay2014frontal}, visual surveillance~\cite{bouchrika2018survey}, and video retrieval~\cite{samangooei2010performing}.

The goal of appearance-based gait recognition is to learn the effective gait pattern representation to distinguish silhouettes of pedestrians with different IDs. Therefore, the representation of gait patterns should be discriminative for different IDs and compact for the same ID.
However, gait recognition is challenging in the following two aspects shown in~\cref{shoutu1}. 

First, the appearance of silhouettes are widely affected by various covariates (e.g., walking speed~\cite{mansur2014gait,tsuji2010silhouette}, carrying conditions~\cite{tao2007general}, and viewpoints~\cite{lu2010uncorrelated}), which lead to large visual differences in the silhouettes of the same ID, i.e., the high intra-class diversity. 
Second, since the silhouettes have no texture and color information, the silhouettes of different IDs could be similar, especially for those under the same covariate settings. This could lead to the high inter-class similarity between the silhouettes of different IDs. 
These two points correspond to two types of pair-wise hard samples in gait recognition: \textit{intra-class hard positives} and \textit{inter-class hard negatives}.
Specifically, the samples of the same ID under different covariates are the intra-class hard positives, and the samples of different IDs under the same covariate are the inter-class hard negatives. And the samples of the remaining cases can be regarded as easy samples.

To solve the hard sample issue, we first investigate the influence of the hard samples on the learning process. In this work, we argue that the main difficulty of learning hard samples in gait recognition is the inconsistent learning objective caused by the different granularities of visual clues between easy/hard samples. As shown in~\cref{shoutu2}, taking the negatives of a specific anchor as an example, for easy negatives, the network could easily distinguish them from the evident and static visual clues like the body shape of silhouettes in a certain frame. However, the network has to capture subtle dynamic motion visual clues from many frames of the gait sequences to distinguish the hard negatives. These huge differences (evident vs. subtle and static vs. dynamic) form the inconsistent objective in the learning process. Besides, a similar phenomenon occurs between easy positives and hard positives.

With the above motivations, we propose a Memory-augmented Progressive Learning network (GaitMPL) with two modules: Dynamic Reweighting Progressive Learning (DRPL) and Global Structure-Aligned Memory bank (GSAM). The core idea of our method is to alleviate the inconsistency of the learning objective caused by hard samples. Specifically, DRPL treats the easy/hard samples separately and works in an easy-to-hard progressive learning fashion, to achieve a more consistent learning objective during training.
Meanwhile, GSAM is further proposed to augment DRPL from the perspective of introducing the memory mechanism to better handle the hard sample issue. Technically, we maintain and model the gait features' distribution of each ID in the memory. Then these features are used in two ways: giving auxiliary supervision signals to distinguish different IDs and constraining the feature variance to compact the representation of the same ID.
Besides, different from the conventional update mechanism of the memory bank, i.e., only parts of elements in memory are updated, we propose a structure-aligned update strategy to maintain the reliability of features in GSAM during training. This strategy preserves the global structure of all IDs' features by updating all the elements in memory simultaneously.

Finally, we test GaitMPL on two commonly used datasets, i.e., CASIA-B~\cite{1699873} and OU-MVLP~\cite{takemura2018multi}. The results show the effectiveness of GaitMPL on gait recognition. Besides, it is worth mentioning that DRPL and GSAM are model-agnostic and can be integrated into most deep gait recognition methods to obtain considerable performance gain, with no extra parameter/computation cost in the inference phase. The main contributions of this work are summarized as follows:

1) We propose a Dynamic Reweighting Progressive Learning module (DRPL) to alleviate the inconsistency of the learning objective caused by the hard samples in gait recognition, which is achieved by reweighting the easy/hard samples in a progressively easy-to-hard learning fashion.

2) We propose a Global Structure-Aligned Memory bank (GSAM), which aims at introducing the global prior information to improve the intra-class compactness and the inter-class separability. This module maintains the centroid feature of each ID and is updated in a novel structure-aligned strategy.

3) Extensive experimental results on two popular datasets demonstrate the effectiveness and generalizability of GaitMPL. The performance gain indicates that the hard sample issue in gait recognition is mainly derived from the clothing condition.

\label{intro}

\section{Related Work}
\subsection{Gait Recognition}
Mainstream gait recognition works focus on utilizing spatial and temporal feature representation to model the gait pattern. Previous works can be mainly reviewed from the following three perspectives: gait representation, network architecture, and optimization objective.

\subsubsection{Gait Representation} 
The representation of gait can be mainly summarized into two parts: silhouette and skeleton. Gait silhouettes are proven to be effective and convenient to express clothing, speed, leg angles, cycle time, etc.~\cite{nieto2015vision, 1561189, wang2010chrono, chen2009frame,sepas2020gait,sepas2020view, rida2019towards}. Skeleton body representation~\cite{gianaria2013gait,nordin2016survey,benouis2016gait, nair2016deep} can be collected with depth-sensing cameras or calculated by pose estimation methods. This representation is robust to appearance changes but more sensitive to occlusion. 

\subsubsection{Network Architecture} 
Plenty of network architectures for gait recognition have emerged these years. For example, set-based methods~\cite{chao2019gaitset, hou2020gait} regard gait as a set consisting of independent frames to extract spatial-temporal information more effectively. Temporal-based methods~\cite{Fan_2020_CVPR,feng2016learning,zhang2020learning,yu2017gaitgan,ariyanto2011model,bodor2009view, 4378964, zhao20063d,liao2020model, 5299188, 6680737,Li_2020_CVPR,makihara2006gait,10.1145/3517199} utilize different approaches to represent temporal information. 3DCNN-based methods~\cite{lin2020gait,wolf2016multi,dou2022metagait,dou2023gaitgci} aim at extracting the spatial-temporal features at the same time.

\subsubsection{Optimization Objective}
Most existing works adopt Triplet loss~\cite{hermans2017defense,zhang2022adaptive}, which compares the anchor with the positives and the negatives. Cross-Entropy loss~\cite{de2005tutorial}, Arcface loss~\cite{deng2019arcface}, Center loss~\cite{wen2016discriminative}, Contrastive loss~\cite{ghojogh2020fisher}, and Euclidean loss~\cite{de1994k} are also the commonly used loss functions. In addition to using a single loss function, there are plenty of works that use hybrid loss functions. GLN~\cite{hou2020gait} adopts the Triplet loss and Cross-Entropy loss in the different training stages. CapsNet~\cite{sepas2020gait} utilizes these two loss functions on its different components of the network.

\subsection{Progressive Learning}

Existing progressive learning can mainly be divided into two categories: curriculum learning~\cite{bengio2009curriculum} and hard example mining (HEM)~\cite{shrivastava2016training}, which represent the ``easier first" and the ``harder first" learning order, respectively.

Curriculum learning has been validated for its effectiveness on several tasks. Zhang et al.~\cite{zhang2018empirical} conduct curriculum learning on the neural machine translation task to alleviate its expensive training cost. Hacohen and  Weinshall~\cite{hacohen2019power} apply curriculum learning to image recognition and prove that curriculum learning effectively modifies the optimization landscape.

Hard example mining (HEM)~\cite{smirnov2018hard, sun2018glancenets,suh2019stochastic,jin2018unsupervised} is also a popular progressive learning paradigm, whose learning order is opposite to curriculum learning. The core of HEM is that the HEM literature assumes the harder samples are more informative~\cite{shrivastava2016training}. Recent works focus on the losses on the samples~\cite{Loshchilov2015OnlineBS,shrivastava2016training,su2023language} or the gradient magnitude~\cite{alain2016variance}.

However, which learning paradigm is more suitable for gait recognition? Chang et al.~\cite{NIPS2017_2f37d101} point out an intuitive criterion: curriculum learning is more suitable for challenging tasks with widespread hard samples. Compared with other vision tasks, it is more difficult to distinguish the differences in silhouette due to the lack of color and texture, ie., there are a large proportion of hard samples in gait recognition. Therefore, the ``easier first" learning order is more suitable for gait recognition. However, the standard curriculum learning achieves its learning order by explicitly controlling the sampling process, which needs to modify the training process. Thus, to improve training efficiency and simplify the training process, this work proposes a batch-level progressive learning strategy, which can be flexibly integrated into most of the existing methods.

% while HEM focuses on accelerating and stabilizing the optimization process in the cleaner dataset.

\begin{figure*}[t]
    \begin{center}
       \includegraphics[width=15cm]{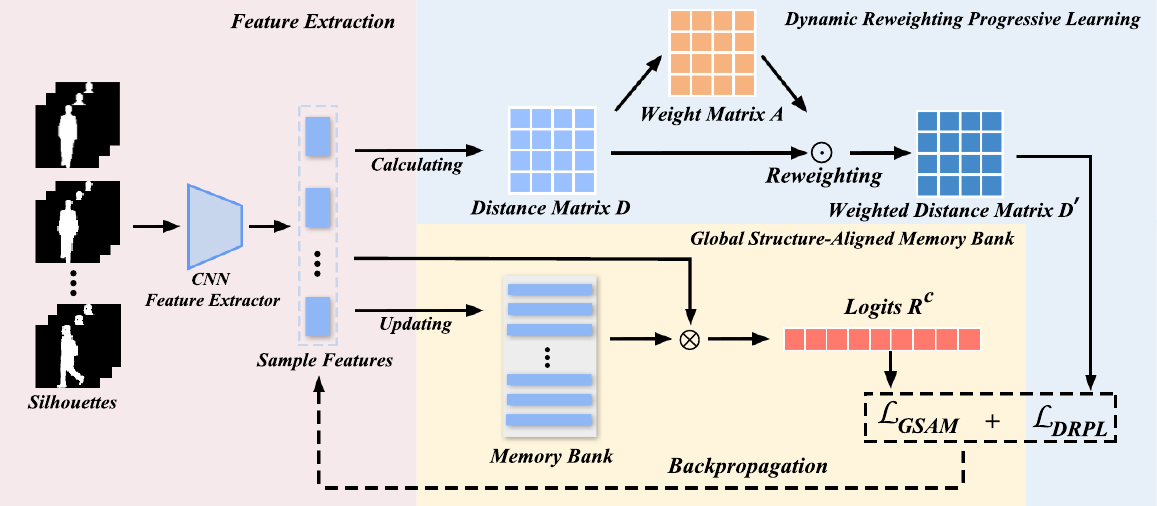}
    \end{center}
    \caption{The overview of GaitMPL. Firstly, the feature extractor extracts the features of input silhouette sequences. Then these features are used to calculate the distance matrix $D$. DRPL aims at generating hardness-based attention weight matrix $A$ for samples and transforming $D$ into the weighted distance matrix $D^{\prime}$. Meanwhile, GSAM introduces a memory mechanism to supervise the training process through logits calculated by the sample features and the memory bank, which could augment progressive learning. In the inference time, only the original distance matrix $D$ is used.}
    \label{fig:pipeline}
 \end{figure*}

\subsection{Learning with Memory Bank}
The memory bank mechanism becomes popular and has been applied to many computer vision tasks. The memory bank mechanism is first proposed by~\cite{wu2018unsupervised} to maintain the feature representation of instances, which avoids calculating the features repeatedly. Several contrastive learning papers~\cite{wu2018unsupervised, he2020momentum, tian2020contrastive, Misra_2020_CVPR, Lai_2020_CVPR} utilize memory bank to compute their loss, which needs plenty of negative samples. For the image retrieval task, there are various works on ReID~\cite{ge2020self, Song_2019_CVPR, Wang_2020_CVPR, li2020joint, Wang_2021_WACV}, which use the memory bank to maintain the feature's scalability, discriminability, and consistency. e.g., Ge et. al.~\cite{ge2020self} use a hybrid memory to encode all available information from the source and target domain for domain adaptive object ReID. Wang et. al.~\cite{Wang_2021_WACV} leverage the memory bank to boost the intra-camera learning performance.

And the memory bank is also usually used as an auxiliary classifier to supervise the training process, to make the inter-class samples more discriminative in the feature space~\cite{Song_2019_CVPR, Wang_2020_CVPR, li2020joint}. In this paper, we take advantage of the property of the memory bank to augment the progressive learning process. Furthermore, considering the feature shift caused by the conventional update strategy, which only updates the IDs present in the current batch, we design a structure-aligned strategy to maintain the reliability of the features in the memory bank.

\section{Method}

In this section, firstly, we give an overview of the proposed GaitMPL framework. Then, we elaborate its two components: Dynamic Reweighting Progressive Learning (DRPL) and Global Structure-Aligned Memory bank (GSAM) in~\cref{GSAM,DRCL}, respectively. At last, we summarize the whole training process of GaitMPL in~\cref{training}.
\label{method}

\subsection{Overview}

Given the silhouette sequences $X\in R^{|\mathcal{B}|\times N_T\times H\times W}$ with the ID labels $Y\in R^{|\mathcal{B}|}$, the goal of gait recognition is to learn the effective gait pattern representation $f$. $|\mathcal{B}|$ is the batch size and $N_T$ is the number of the frame of each sequence. $H$ and $W$ are the height and width of the silhouette, respectively. A common learning pipeline is to use a feature extractor $F$ to obtain the gait features $f\in R^{|\mathcal{B}|\times N_{dim}}$ with dimension $N_{dim}$ as follows:
\begin{equation}
    f_i= F(x_i), \quad s.t. \ \  x_i \in X.
\end{equation}

Then the feature extractor is optimized by its objective function. In this work, we propose a new learning pipeline with two components, i.e., DRPL and GSAM. This pipeline aims at addressing the intra-class and inter-class hard sample issue, which is usually left out by the conventional pipelines. Moreover, DRPL and GSAM are model-agnostic and can be applied to various deep models. The overall illustration of our pipeline is shown in~\cref{fig:pipeline}.

\subsection{Dynamic Reweighting Progressive Learning} 
\label{DRCL}

As discussed in the introduction, gait recognition is challenging due to the various covariates and the properties of silhouettes (textureless and colorless), which correspond to two kinds of hard samples. And the different granularities of visual clues between easy/hard samples result in inconsistent learning objective. To tackle this hard sample issue, we propose the DRPL module. The core idea of DRPL is to alleviate the inconsistent learning objective with an easy-to-hard progressive learning strategy.
Specifically, DRPL treats the easy/hard samples separately through the control of the attention weight distribution of samples. Further, considering the subtle and dynamic visual clues of hard samples are harder to learn than the evident and static visual clues of easy samples, DRPL conducts progressive learning to pay more attention to hard samples with the training going on. By doing this, the learning objective is more consistent at each training stage for gait recognition.
To achieve easy-to-hard progressive learning, DRPL has two parts: hardness-based attention to identify the hard samples and a progressive learning arrangement to determine the learning order.

 \begin{figure}[t]
    \begin{center}
       \includegraphics[width=8cm]{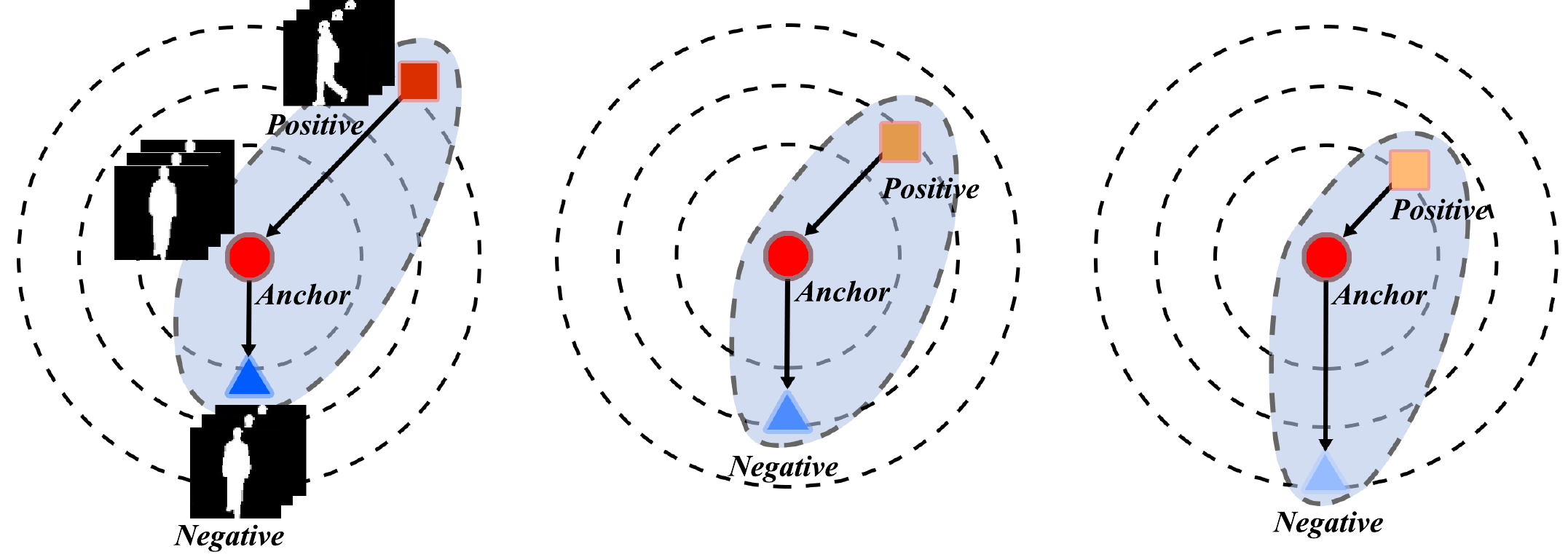}
    \end{center}
    \caption{The illustration of the sample hardness. The harder samples are marked with a darker color. For positive samples, the hard samples have a larger feature distance with the anchor. For negative samples, the hard samples have a smaller feature distance with the anchor.}
    \label{drcla}
 \end{figure}

\subsubsection{Hardness-Based Attention} 
The key to progressive learning is how to define the hardness of samples. Considering that the data in a batch is randomly selected from the whole data distribution, we could directly measure the relative hardness of samples based on the distance from other samples in the same batch. Specifically, as shown in~\cref{drcla}, we model the relative hardness of samples according to their feature distance from the anchor $x_i$. In general, the positives $x \in S_{x_i}^+$ with large distance or negatives $x \in S_{x_i}^-$ with small distance compared with the anchor sample are regarded as the hard samples, which should have large hardness values.

To associate the hardness of samples with the learning process, we calculate the hardness-based attention weight for each sample according to their relative hardness at the batch level. Specifically, the calculation for this hardness-based attention weights of the samples is defined as follows:
\begin{equation}
W_{x_i,x,t}=
    \begin{cases}
        \sigma(\exp{(\frac{d(F(x_i),F(x))}{\delta_{t}})})\ \ \ \ \ \ & if\  x \in S_{x_i}^+\\
    
        \sigma(\exp{(-\frac{d(F(x_i),F(x))}{\delta_{t}})}) \ \  &otherwise,\\
    \end{cases}
\label{metric}
\end{equation}
where $d(\cdot,\ \cdot)$ is the function to calculate Euclidean distance and $\sigma(\cdot)$ is the Softmax function to normalize the attention weights among positives or negatives, which could measure the relative hardness in the same batch.
Notably, $\delta_{t}$ is a factor to control the softness of the attention weight distribution.
In this way, the samples in the same batch could obtain different attention weights according to their relative hardness adaptively to achieve treating easy/hard samples separately.

\subsubsection{Progressive Arrangement} 
\begin{figure}[t]
    \begin{center}
       \includegraphics[width=8cm]{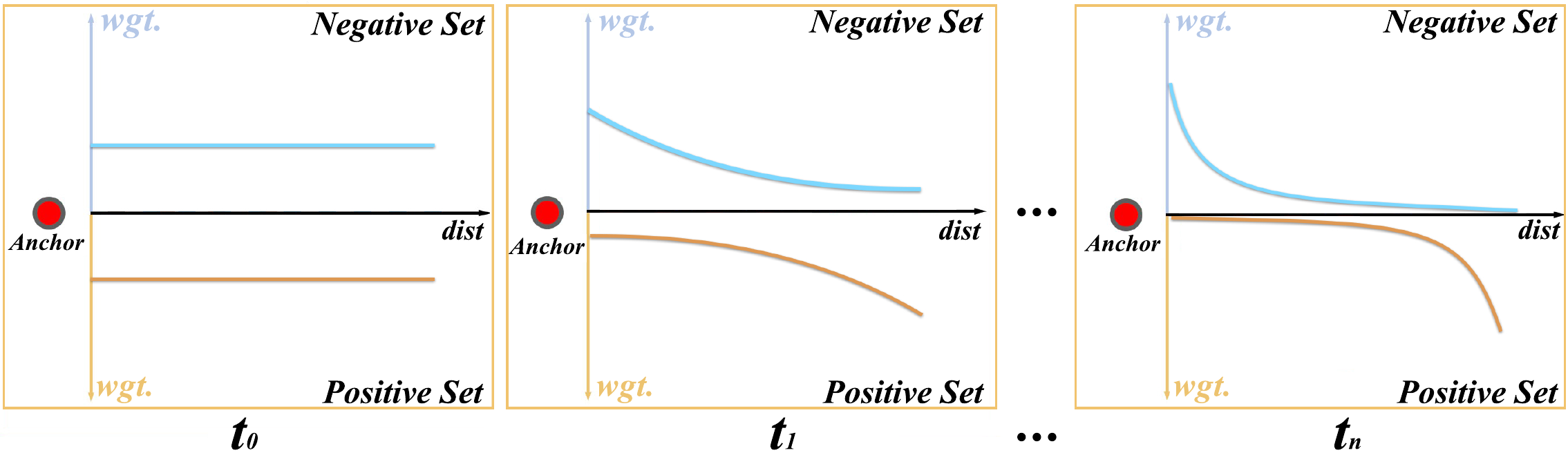}
    \end{center}
    \caption{The change of attention weight distribution on negatives (blue curve) and positives (brown curve). With the progress of the training process, DRPL makes the distribution of sample weights sharper progressively, which pays growing attention to the hard samples.}
    \label{drclb}
 \end{figure}

With the hardness-based attention calculation proposed in~\cref{metric}, we could arrange the learning order flexibly. To accomplish progressive learning from easy to hard during the training process, we design an explicit method to control the attention weight distribution of samples, aiming at paying growing attention to the hard samples with the training going on. Specifically, we use the dynamic progressive factor $\delta_t$ described in~\cref{metric} to control the learning process, which is formulated as follows:

\begin{equation}
    \delta_t=\delta_{min} + \frac{1}{2}(1+cos(\frac{t}{2T}\pi))\cdot \epsilon,
    \label{curriculum}
\end{equation}
where $\epsilon$ is a hyperparameter that corresponds to the change speed of the progressive factor and $\delta_{min}$ is another hyperparameter to ensure the numerical stability. Moreover, $t$ and $T$ denote the current iteration and the total iterations, respectively. As shown in~\cref{drclb}, by reducing the progressive factor $\delta_t$, we make the attention weight distribution sharper progressively. In this way, our pipeline establishes a batch-level easy-to-hard progressive learning paradigm, which treats the easy/hard samples separately and pays growing attention to hard samples, to alleviate the issue of the inconsistent learning objective.

\subsubsection{Objective Function Construction} 
With the adaptive attention weight distribution on samples, we revise the feature distance as follows:
\begin{equation}
    d^{\prime}_t(F(x_i), F(x)) = W_{x_i,x,t} \cdot d(F(x_i), F(x)).
    % \begin{cases}
    %     d'_t(F(x_a^i), F(x_p^i)) =W_{x_a^i,x_p^j,t} \cdot d(F(x_a^i), F(x_p^i))\\
    %     d'_t(F(x_a^i), F(x_n^j)) =W_{x_a^i,x_n^j,t} \cdot d(F(x_a^i), F(x_n^j))
    % \end{cases}.
    \label{reweighting}
\end{equation}

We design a flexible objective function by replacing the distance of the conventional triplet loss with our weighted feature distance in a batch-hard-like (BH-like) formulation:

\begin{equation}
    \begin{small}
    \begin{aligned}
    % \mathcal{L}_{DRCL}(f_\theta;X)=\overbrace{\sum\limits_{i=1}^P\sum\limits_{a=1}^K}^{\textnormal{all anchors}}[m+&\overbrace{\max\limits_{p=1\dots K}d^{'}(f_{\theta}(x_a^i),f_{\theta}(x_p^i))}^{\textnormal{positive}}-\\
    % &\underbrace{\min_{\substack{j=1\dots P\\n=1\dots K\\j\neq i}}d^{'}(f_{\theta}(x_a^i),f_{\theta}(x_n^j))}_{\textnormal{negative}}]_+
    \mathcal{L}_{DRPL-BH}=&\sum\limits_{i=1}^P\sum\limits_{a=1}^K[m+\overbrace{\frac{1}{K}\sum_{p=1}^K{d^{\prime}_t(F(x_{i,a}),F(x_{i,p}))}}^{\textnormal{positive}}-\\
    &\underbrace{\frac{1}{K(P-1)}\sum_{\substack{j=1\dots P,\ j\neq i,\\n=1\dots K}}{d^{\prime}_t(F(x_{i,a}),F(x_{j,n}))}}_{\textnormal{negative}}]_+,
    \end{aligned}
\end{small}
\label{drplbh}
\end{equation}
where $P$ and $K$ denote the number of IDs and sequences for each ID in a batch, respectively. Besides, $x_{i,a}$ means the anchor sample of ID $i$, $x_{i, p}$ represents the positive sample of ID $i$, and $x_{j, n}$ is the negative sample of ID $j$. 

To ensure the smoothness of optimization, we build the final DRPL loss using both the standard batch-all (BA) triplet loss~\cite{hermans2017defense}  and our batch-hard-like DRPL loss as follows:
\begin{equation}
    \begin{small}
    \begin{aligned}
    % \mathcal{L}_{DRCL}(f_\theta;X)=\overbrace{\sum\limits_{i=1}^P\sum\limits_{a=1}^K}^{\textnormal{all anchors}}[m+&\overbrace{\max\limits_{p=1\dots K}d^{'}(f_{\theta}(x_a^i),f_{\theta}(x_p^i))}^{\textnormal{positive}}-\\
    % &\underbrace{\min_{\substack{j=1\dots P\\n=1\dots K\\j\neq i}}d^{'}(f_{\theta}(x_a^i),f_{\theta}(x_n^j))}_{\textnormal{negative}}]_+
    \mathcal{L}_{Triplet-BA}=&\sum\limits_{i=1}^P\sum\limits_{a=1}^K\sum\limits_{\substack{p=1\\p\neq a}}^{K}\sum\limits_{\substack{j=1\\j\neq i}}^P\sum\limits_{n=1}^{K}[m+{d(F(x_{i,a}),F(x_{i,p}))}-\\
    &{d(F(x_{i,a}),F(x_{j,n}))}]_+,
    \end{aligned}
\end{small}
\end{equation}

\begin{equation}
    \mathcal{L}_{DRPL}=\mathcal{L}_{Triplet-BA}+s_t\mathcal{L}_{DRPL-BH},
    \label{loss_drpl}
\end{equation}
where $s_t$ is a smoothness factor related to the training iteration $t$, which can improve the stability of batch-hard-like loss. There are many alternative implementations, and we use a sine style function. And the effect of the smoothness factor is discussed in~\cref{ablation_exp}.

\subsection{Global Structure-Aligned Memory Bank}
\label{GSAM}

To augment the learning process of DRPL mentioned above, we propose a GSAM module, which aims at using the global prior information to supervise the training process through the memory mechanism. Specifically, there are two functionalities of GSAM. First, GSAM is adopted as an auxiliary classifier to make the inter-class features more discriminative. Second, GSAM also constrains the intra-class variance to make the intra-class features more compact. Different from the common memory mechanism using the conventional momentum update strategy, GSAM uses a structure-aligned update strategy, which can preserve the global structure of the features to maintain the reliability of features in GSAM. 
Moreover, GSAM has a more compact representation for the elements in memory and can reduce the parameter cost.
In this part, we introduce GSAM from three aspects: how to design the elements in GSAM, how to update the elements in GSAM, and how to supervise the training process with GSAM.

\subsubsection{The Memory Elements of GSAM}
The elements maintained in the memory bank are very important because the global prior information should be discriminative for representing the IDs and comprehensively model the patterns of pedestrians, which should cover all covariates information. More importantly, due to the resource limit, it is infeasible to maintain the features of gait silhouette sequences under all covariates. Therefore, we propose to use the feature centroids of each ID to represent the ID-level features, which are formulated as:
\begin{equation}
    \mathcal{M}[i] = \frac{\sum_{s\in S_i}{F(s)}}{N_s},
    \label{memory_define}
\end{equation}
where the element of the ID $i$ of GSAM $\mathcal{M}\in R^{N_{id}\times N_{dim}}$ is by calculating the centroid features of all the corresponding sequences $S_i$. $N_{id}$, $N_{dim}$, and $N_s$ is the number of IDs, the feature dimension, and the number of the sequences of each ID, respectively. The features are all extracted by the feature extractor $F$. Therefore, GSAM maintains robust ID-level features by covering various covariates of the same ID. The advantages of this representation lie in two aspects: \romannumeral1) it can cover the information of various covariates, therefore it is robust to the cross-viewpoint or cross-walking-condition cases; \romannumeral2) it avoids the huge parameter cost by fusing plenty of sequences into one representation.

\subsubsection{The Update Strategy of GSAM} 
\label{update_section}

\begin{figure}[t]
    \begin{center}
       \includegraphics[width=7cm]{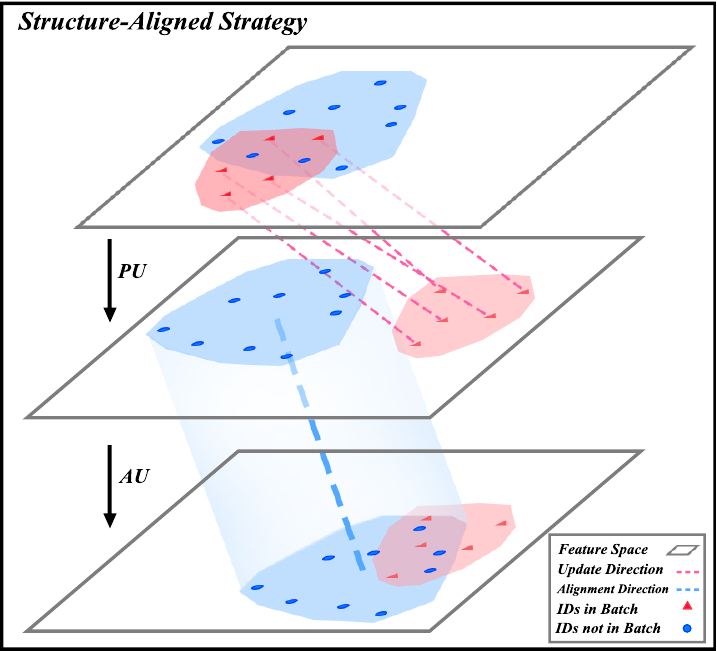}
    \end{center}
    \caption{The illustration of structure-aligned strategy in the feature space. \romannumeral1) Only making the IDs Present in the current batch Updated (PU) could cause the feature shift between the present IDs and the absent IDs in the feature space; \romannumeral2) Structure-Aligned strategy not only adopts the PU strategy but also makes the IDs Absent in the current batch Updated (AU) towards the same alignment direction, the relative structure of the feature space can be aligned to maintain the reliability of features in GSAM.}
    \label{sa}
 \end{figure}
The update strategy is crucial to the GSAM. Most previous methods~\cite{Song_2019_CVPR,ge2020self, Wang_2020_CVPR,li2020joint} follow the conventional momentum updating strategy, which only makes the IDs Present in the current batch Updated (PU). However, this strategy causes the feature shift between the present IDs and absent IDs in the feature space. As shown in~\cref{sa}, if only adopting the PU strategy, there is an inevitable feature shift compared with the absent IDs in the feature space. This phenomenon reduces the reliability of global prior information in GSAM, which would confuse the training process and cannot guide the architecture towards the correct optimization direction.

To ensure the reliability of the features in GSAM, we design a structure-aligned strategy to preserve the relative structural position of the features during optimization, which not only makes the IDs Present in the current batch Updated (PU) but also makes the IDs Absent in the current batch Updated (AU). And the effects of structure-aligned strategy on the feature space are shown in~\cref{sa}. It maintains the relative structural position of all IDs during training, making the memory elements effectively in representing global prior information. And this strategy can be formulated as follows:

\begin{footnotesize}
    \begin{equation}
        \begin{aligned}
        \mathcal{M}[i]^t=
           \begin{cases}
            \frac{(1-\alpha)}{|\mathcal{B}|}\sum\limits_{\substack{b\in \mathcal{B}, y_b=i}}F(b) + \alpha\mathcal{M}[i]^{t-1}\,\,\,\,\,\,\,\,\,\,\,\,\,\,\,\,\,\,\,\,\,\,\,\,\,\,\,\,\,\,\,\,\,\,\,\, \mbox{if $i \in$  $y_\mathcal{B}$}\\
            
           (1-\beta)(\frac{1}{|\mathcal{B}|}\sum\limits_{b\in \mathcal{B}}F(b)-\mathcal{M}[i]^{t-1}) +\beta\mathcal{M}[i]^{t-1}\,\,otherwise.
           \end{cases}
        \end{aligned}
       \label{update}
      \end{equation}
    \end{footnotesize}

    \cref{update} shows the memory update strategy for the ID $i$. And the label set $y_\mathcal{B}$ is composed of the IDs present in the current batch $\mathcal{B}$ under the iteration $t$. $|\mathcal{B}|$ is the batch size. For the IDs present in the current batch, the corresponding elements in GSAM are updated under the momentum update strategy with the parameter $\alpha$. Meanwhile, for the IDs absent in the current batch, the corresponding elements are updated towards the alignment direction of averaged movement of the IDs in the current batch with momentum $\beta$.

\subsubsection{The Supervision under GSAM} In this part, we show how to utilize GSAM during training. GSAM can be used as global prior information to supervise the training process through a typical Cross-Entropy loss. Besides, to further ensure the compactness in the feature space, GSAM can also be used to constrain the intra-class variance. The loss function that corresponds to GSAM is defined as:

\begin{equation}
\begin{small}
    \begin{aligned}
        \mathcal{L}_{GSAM}={\mathcal{L}_{CE}(F(X)\mathcal{M}^T/\tau, Y)}+(F(X)-\overline{F(X)})^2,
    \end{aligned}
\end{small}
   \label{lmemory}
  \end{equation}
where the operation betweem $F(x)$ and $\mathcal{M}$ is matrix multiplication. $\tau$ denotes the temperature to balance the feature distribution, and $\overline{F(x)}$ denotes the averaged feature along batch dimension.

To emphasize, in the training phase, the parameter and computation cost introduced by GSAM increases linearly with the number of IDs. In the inference phase, GSAM is not involved, thus it does not introduce any extra cost.

\subsection{The Whole Training Process}
 Since the feature extractor is not reliable at the early training stages, the memory elements at the early stages may be unstable to represent the global prior information. Therefore, the weight for $\mathcal{L}_{GSAM}$ should be smaller at the early training stages to stabilize the training process, which requires that the fusion factor $\gamma_t$ should be monotonically increasing with the increase of training iterations. The total loss function of our method is formulated as:
\begin{equation}
    \mathcal{L}_{total}= \mathcal{L}_{DRPL}+\gamma_t\mathcal{L}_{GSAM},
\label{total}
\end{equation}
where $\gamma_t=sin(\frac{\pi t}{2T})$ is a fusion factor and it focuses on smoothing the training process. $t$ is the current iteration and $T$ is the total iterations. And the optimization process of GaitMPL is summarized in Alg.~\ref{alg:training}.

% and will be discussed in~\cref{ablation_exp}.

\begin{algorithm2e}[h]
    \caption{Training Process of GaitMPL}
    \scriptsize
    \label{alg:training}
    \SetKwInput{Kwinit}{Init}{}{}
    \SetKwInput{kwInput}{Input}{}{}
    \SetKwInput{kwOutput}{Output}{}{}
    \kwInput{Training data $X = \{ x_1, x_2,..., x_N\}$, ID set $\mathcal{I}$, feature extractor $F$, memory bank $\mathcal{M}$, max iteration number $T$.} 
    % \kwOutput{Optimized parameter $\vartheta$}
    \kwOutput{Trained feature extractor $F$.}
    % \Kwinit{Feature encoder $F$ initialized by $\vartheta$; \\
    %         \qquad Compute $\mathcal{M} = \{m_1,...,m_N \}$ based on ~\cref{memory_define} ; \\} 
    \For{i {\rm in $\mathcal{I}$}}
    {
        // {\color{gray}Initialize memory bank} \\
        $\mathcal{M}[i]\gets \frac{1}{N_s}\sum\limits_{\substack{b \in X, y_b=i}}F(b)$ ;
    }
    \For{iter = 1,..., T}{
        // {\color{gray}Calculate the features $f = (f_i)_{|\mathcal{B}|}$ of samples in current batch $\mathcal{B}$}\\
        $f_i \gets F(x_i)\,\,\,\,\,\,\, \mathrm{s.t.}\,\,\,\,\,\, x_i \in \mathcal{B}$;\\
        Update $\mathcal{M}$ based on~\cref{update};\\
        // {\color{gray}Calculate the distance matrix $D = (D_{ij})_{|\mathcal{B}| \times |\mathcal{B}|}$}\\
        $D_{ij} \gets d(f_i, f_j)$;\\
        Reweight $D$ into $D^{\prime}$ based on~\cref{reweighting};\\
        Compute the  $\mathcal{L}_{DRPL}$ with $D^{\prime}$ based on~\cref{loss_drpl};\\
        Compute the $\mathcal{L}_{GSAM}$ with $\mathcal{M}$ based on~\cref{lmemory};\\
        Compute the $\mathcal{L}_{Total}$ based on~\cref{total};\\
        Backpropagation and update feature extractor $F$;
    }
    Return $F$;
%       }
\end{algorithm2e}
\label{training}

% \vspace{-1em}

\begin{table*}[t]
    \centering
    \scriptsize
    \caption{The averaged rank-1 accuracies on \textbf{casia-b}, identical views are excluded. We mark the methods reproduced by ourselves with $\dagger$. Notably, GLN uses a higher resolution of $128\times 88$, while the resolution of the rest methods is $64\times 44$. \\GaitMPL adopts MT3D and GaitGL as the feature extractor of GaitMPL(M) and GaitMPL(G), respectively.}
    \scalebox{1}[0.85]{
    \begin{tabular}{c|c|ccccccccccc|c}
    \toprule
    \multicolumn{2}{l|}{Gallery NM\#1-4}  & \multicolumn{11}{c|}{0°-180°}                                                                                                                                                                                                                                                                                                                                                                                                       &                                       \\ \cline{1-13}
    \multicolumn{2}{l|}{Probe}            & \multicolumn{1}{l}{0°}                & \multicolumn{1}{l}{18°}               & \multicolumn{1}{l}{36°}               & \multicolumn{1}{l}{54°}               & \multicolumn{1}{l}{72°}               & \multicolumn{1}{l}{90°}               & \multicolumn{1}{l}{108°}              & \multicolumn{1}{l}{126°}              & \multicolumn{1}{l}{144°}              & \multicolumn{1}{l}{162°}              & \multicolumn{1}{l|}{180°}    & \multirow{-2}{*}{Mean}                \\ \hline
                               & CNN-LB   & 82.6                                 & 90.3                                 & 96.1                                 & 94.3                                 & 90.1                                 & 87.4                                 & 89.9                                 & 94.0                                 & 94.7                                 & 91.3                                 & 78.5                        & 89.9                                  \\ 
                               & GaitSet $\dagger$  & 86.7                                 & 97.1                                 & 98.4                                 & 96.0                                 & 94.0                                 & 91.8                                 & 95.0                                 & 96.8                                 & 97.4                                 & 96.5                                 & 86.0                        & 94.2                                  \\  
                               & GaitNet  & 91.2                                 & 92.0                                 & 90.5                                 & 95.6                                 & 86.9                                 & 92.6                                 & 93.5                                 & 96.0                                 & 90.9                                 & 88.8                                 & 89.0                        & 91.6                                  \\  
                               & GaitPart $\dagger$ & 92.6                                 &  98.2          & 98.7         & 98.1         &95.3          &  93.4         &  96.4          &  98.0          & 99.0          &  97.8         &  89.1 & 96.1                                  \\
                               & GLN      &  93.2          &  99.3 &  99.5 &  98.7 &  96.1 & 95.6 &  97.2 &  98.1 & 99.3 &  98.6 &  90.1 & 96.9 \\  
                               & MT3D $\dagger$    & 95.4 & 97.4                                 &  99.0            &  97.1         &  95.5          & 95.0                                   & 96.6         & 98.9                                 & 98.6                                 & 98.3                                 & 90.9               &  96.6          \\ 
                               & DynamicGait  & 90.6                                 & 97.9                                & 98.5                                 & 97.6                                 & 93.9                                 & 89.7                                 & 94.4                                 & 98.1                                 & 97.8                                 & 96.2                                 & 87.9                       & 94.8                                  \\  
                               &3DLocal  & 96.0 & 99.0 & 99.5 & 98.9 & 97.1 & 94.2 & 96.3 & 99.0 & 98.8 & 98.5 & 95.2 & 97.5                  \\
                               &CSTL    & 97.2 & 99.0 & 99.2 & 98.1 & 96.2 & 95.5 & 97.7 & 98.7 & 99.2 & 98.9 & 96.5& 97.8                  \\
                             &  GaitGL   &      96.0 &98.3& 99.0& 97.9& 96.9 &95.4& 97.0& 98.9& 99.3& 98.8& 94.0& 97.4    \\
    & \textbf{GaitMPL(M)}     & 93.1                                 & 96.8                                 & 98.8                                 & 96.1                                 & 94.9                                 & 92.9                                 & 95.4                                 & 98.3                                 & 98.3                                 & 98.0                                   & 88.1                        & 95.5                                  \\ 
    \multirow{-11}{*}{NM \#5-6} & \textbf{GaitMPL(G)}     & 97.2&98.3&99.5&98.3&96.5&95.2&97.2&98.5&99.1&98.5&93.8&97.5         \\ \hline
                               & CNN-LB   & 64.2                                 & 80.6                                 & 82.7                                 & 76.9                                 & 64.8                                 & 63.1                                 & 68.0                                 & 76.9                                 & 82.2                                 & 75.4                                 & 61.3                        & 72.4                                  \\  
                               & GaitSet $\dagger$  & 81.0                                 & 90.6                                 & 94.5                                 & 92.8                                 & 87.3                                 & 82.2                                 & 85.9                                 & 90.9                                 & 92.8                                 & 92.4                                 & 79.5                        & 88.2                                  \\  
                               & GaitNet  & 83.0                                 & 87.8                                 & 88.3                                 & 93.3                                 & 82.6                                 & 74.8                                 & 89.5                                 & 91.0                                 & 86.1                                 & 81.2                                 & 85.6                        & 85.7                                  \\ 
                               & GaitPart $\dagger$ & 88.6                                 & 94.8                                 & 96.2                                 & 95.1                                 & 90.8                                 & 85.1                                 & 89.5                                 & 95.2                                 & 96.9                                 & 95.8                                 & 84.7                        & 92.1                                  \\  
                               & GLN      & 91.1                                 & 97.7                        & 97.8                        & 95.2                       & 92.5                                 & 91.2                        & 92.4                        & 96.0                        & 97.5                                 & 94.9                                & 88.1               & 94.0                       \\ 
                               & MT3D $\dagger$    & 91.2                       & 95.8                                 & 95.5                                 & 93.7                                 & 92.1                                 & 87.7                                 & 92.0                                 & 95.4                                 & 96.8                                 & 96.5                      & 86.0                        & 92.9                                  \\ 
                               & DynamicGait  & 84.0                                 & 91.8                                & 94.8                                 & 92.7                                 & 85.7                                & 82.1                                 & 86.7                                 & 91.7                                 & 94.4                                 & 91.9                                 & 81.3                       & 88.8                                  \\  
                               & 3DLocal  & 92.9 & 95.9 & 97.8 & 96.2 & 93.0 & 87.8 & 92.7 & 96.3 & 97.9 & 98.0 & 88.5 & 94.3    \\
                               & CSTL     & 91.7 & 96.5 & 97.0 & 95.4 & 90.9 & 88.0 & 91.5 & 95.8 & 97.0 & 95.5 & 90.3 & 93.6  \\
                               & GaitGL   &    92.6 &96.6& 96.8 &95.5& 93.5& 89.3& 92.2& 96.5& 98.2& 96.9& 91.5& 94.5  \\
                               
    \multirow{-9}{*}{BG \#1-2} & \textbf{GaitMPL(M)}    & 89.1                                 & 94.6                                 & 96.1                                 & 94.1                                & 93.5                       & 88.7                                 & 91.9                                 & 95.7                                 & 98.0                        & 95.3                                & 84.3                        & 92.9                                  \\ 
     & \textbf{GaitMPL(G)}     & 92.3&96.8&95.8&95.4&93.8&89.5&92.5&95.8&98.4&97.1&91.8&94.5         \\ \hline
                               & CNN-LB   & 37.7                                 & 57.2                                 & 66.6                                 & 61.1                                 & 55.2                                 & 54.6                                 & 55.2                                 & 59.1                                 & 58.9                                 & 48.8                                 & 39.4                        & 54.0                                  \\ 
                               & GaitSet $\dagger$ & 59.9                                & 77.7                                 & 83.2                                 & 78.2                                 & 73.8                                 & 68.7                                 & 69.4                                 & 75.7                                 & 75.4                                 & 73.7                                 & 54.3                        & 71.8                                  \\  
                               & GaitNet  & 42.1                                 & 58.2                                 & 65.1                                 & 70.7                                 & 68.0                                 & 70.6                                 & 65.3                                 & 69.4                                 & 51.5                                 & 50.1                                 & 36.6                        & 58.9                                  \\ 
                               & GaitPart $\dagger$ & 69.3                                 & 84.7                                 & 85.4                                 & 83.5                                 & 78.1                                 & 72.5                                 & 78.9                                 & 81.5                                 & 83.9                                 & 78.6                                 & 63.4                        & 78.2                                  \\  
                               & GLN      & 70.6                                 & 82.4                                 & 85.2                                 & 82.7                                 & 79.2                                 & 76.4                                 & 76.2                                 & 78.9                                 & 77.9                                 & 78.7                                 & 64.3                        & 77.5                                  \\ 
                               & MT3D $\dagger$    & 78.9                                 & 88.0                                   & 89.8                                 & 84.5                                 & 80.7                                 & 75.9                                 & 81.6                                 & 85.2                                 & 86.3                                 & 83.0                                   & 70.1                        & 82.2                                  \\  
                               & DynamicGait  & 75.9                                 & 87.5                                & 90.6                                 & 85.3                                & 81.5                                & 76.5                                 & 81.1                                 & 86.5                                 & 85.5                                 & 82.1                                 & 66.8                      & 81.8                                  \\  
                               & 3DLocal  & 78.2 & 90.2 & 92.0 & 87.1 & 83.0 & 76.8 & 83.1 & 86.6 & 86.8 & 84.1 & 70.9 & 83.7  \\
                               & CSTL     & 78.1 & 89.4 & 91.6 & 86.6 & 82.1 & 79.9 & 81.8 & 86.3 & 88.7 & 86.6 & 75.3 & 84.2  \\
                               & GaitGL   &     76.6& 90.0& 90.3& 87.1& 84.5& 79.0& 84.1& 87.0& 87.3& 84.4& 69.5& 83.6                   \\
    & \textbf{GaitMPL(M)}     & 83.1                       & 93.5                        & 95.4                        & 91.2                        & 88.3                        & 82.5                        & 87.5                     & 91.6                        & 91.2                        & 88.3                        & 74.4               & 87.9                         \\ 
    \multirow{-11}{*}{CL \#1-2}&\textbf{GaitMPL(G)} & 83.8&85.5&96.3&93.0&89.8&83.2&87.3&92.3&91.4&88.7&76.7&88.0        \\  \hline

                               & CNN-LB   & 61.5                                 & 76.0                                 & 81.8                                 & 77.4                                 & 70.0                                 & 68.4                                 & 71.0                                 & 76.7                                 & 78.6                                 & 71.8                                 & 59.7                        & 72.1                                  \\
                               & GaitSet $\dagger$ & 75.9                                 & 88.5                                 & 92.0                                 & 89.0                                 & 85.0                                 & 80.9                                 & 83.4                                 & 87.8                                 & 88.5                                 & 87.5                                 & 73.3                        & 84.7                                  \\
                               & GaitNet  & 72.1                                 & 79.3                                 & 81.3                                 & 86.5                                 & 79.2                                 & 79.3                                 & 82.8                                 & 85.5                                 & 76.2                                 & 73.4                                 & 70.4                        & 78.7                                  \\
                               & GaitPart $\dagger$& 83.5                                 & 92.6                                 & 93.4                                 & 92.2                                 & 88.1                                 & 83.7                                 & 88.3                                 & 91.6                                 & 93.3                                 & 90.7                                 & 79.1                        & 88.8                                  \\
                               & GLN      & 85.0                                 & 93.1                                 & 94.2                                 & 92.2                                 & 89.3                                 & 87.7                                 & 88.6                                 & 91.0                                 & 91.6                                 & 90.8                                 & 80.8                        & 89.5                                  \\
                               & MT3D $\dagger$    & 88.5                        & 93.7                                 & 94.8                                 & 91.8                                 & 89.4                                 & 86.2                                 & 90.1                                 & 93.2                                 & 93.9                                 & 92.6                                 & 82.3                        & 90.6                                  \\
                               & DynamicGait  & 83.5                                 & 92.4                                & 94.6                                 & 91.8                                 & 87.0                                & 82.8                                 & 87.4                                 & 92.1                                 & 92.6                                 & 90.1                                 & 78.7                       & 88.5                                  \\  
                               &3DLocal&89.0&95.0&96.4&94.1&91.0&86.3&90.7&94.0&94.5&93.5&84.9&91.8 \\
                               &CSTL&89.0&95.0&95.9&93.4&89.7&87.8&90.3&93.6&95.0&93.7&87.4&91.9\\
                               &GaitGL& 88.4&95.0&95.4&93.5&91.6&87.9&91.1&94.1&94.9&93.4&85.0&91.8\\

    & \textbf{GaitMPL(M)}     & \textbf{88.4}                                 & \textbf{95.0}                        & \textbf{96.8}                        & \textbf{93.8}                        & \textbf{92.2}                        & \textbf{88.0}                        & \textbf{91.6}                        & \textbf{95.2}                        & \textbf{95.8}                        & \textbf{93.9}                        & \textbf{82.3}               & \textbf{92.1}   \\
    \multirow{-11}{*}{Mean}  & \textbf{GaitMPL(G)} &   \textbf{91.1}&\textbf{93.5}&\textbf{97.2}&\textbf{95.6}&\textbf{93.4}&\textbf{89.3}&\textbf{92.3}&\textbf{95.5}&\textbf{96.3}&\textbf{94.8}&\textbf{87.4}&\textbf{93.3}   \\
    \bottomrule
    \end{tabular}}
    \label{mainresults}
    \end{table*}

\section{Experiments}
\subsection{Datasets and Evaluation Protocols}
\subsubsection{CASIA-B~\cite{1699873}} 
This dataset is the most commonly used dataset, which is composed of 124 IDs. Each ID contains 11 viewpoints uniformly distributed in [0$^{\circ}$, 180$^{\circ}$], and each viewpoint contains 10 sequences, i.e., 6 NM (Normal), 2 BG (with Bag), and 2 CL (Clothing). Thus, this dataset contains 124 $\times$ 11 $\times$ (6+2+2) = 13640 sequences. For dataset partition, there are several mainstream protocols~\cite{Zhang}. For justification, our experiments use the popular protocol proposed by ~\cite{Wu2017}. We take the first 74 IDs for training and the rest 50 IDs for testing. For evaluation, in each ID, the first 4 sequences of NM condition (NM \#1-4) are regarded as the gallery, and the remaining 6 sequences are regarded as the probe.

\subsubsection{OU-MVLP~\cite{takemura2018multi}} 
This dataset is the largest public gait dataset, which is composed of 10307 IDs (5153 IDs for training and 5154 IDs for testing). Besides, all instances are regarded as NM. Each ID has 14 viewpoints, which are more diverse than the viewpoints in CASIA-B, and they are uniformly distributed between [0$^{\circ}$, 90$^{\circ}$] and [180$^{\circ}$, 270$^{\circ}$] at an interval of 15$^{\circ}$. For evaluation, the first sequence of each ID is treated as the gallery while the rest is the probe.

\subsection{Implementation Details}

We adopt MT3D~\cite{lin2020gait} and GaitGL~\cite{Lin_2021_ICCV} as the feature extractor of GaitMPL(M) and GaitMPL(G), respectively. Since DRPL and GSAM are model-agnostic to be applied to any method, thus all network configuration of GaitMPL is the same as the selected feature extractor, such as architecture, feature dimension, resolution, optimizer, and batch size.

We adopt the same hyperparameters on all methods and datasets. In DRPL, $\delta_{min}$ is set to 0.1 and $\epsilon$ is set to 1.4. The smoothness factor in~\cref{loss_drpl} is adopted as $s_t=sin(\frac{\pi t}{2T})$, which is a function of the current iteration $t$ and the total iterations $T$. In GSAM, the memory bank size is set to 74 and 5153 on CASIA-B and OU-MVLP, respectively. The temperature $\tau$ is set to 0.1 and the momentum $\alpha$, $\beta$ are both set to 0.9. It's worth mentioning that we reproduce the results for GaitSet, GaitPart, and MT3D with the data augmentation proposed in~\cite{wu2020condition}, which are slightly higher than their original results~\cite{chao2019gaitset,Fan_2020_CVPR,lin2020gait}.
\label{setting}

\subsection{Comparison with State-of-the-Art Methods}

\subsubsection{CASIA-B} 
To verify the ability of GaitMPL under cross-walking-condition and cross-view scenarios, the comparison is conducted on CASIA-B. We compare the performance of GaitMPL with several state-of-the-art methods, including CNN-LB~\cite{wu2016comprehensive}, GaitSet~\cite{chao2019gaitset}, GaitPart~\cite{Fan_2020_CVPR}, GLN~\cite{hou2020gait}, MT3D~\cite{lin2020gait}, DynamicGait~\cite{wu2020condition}, CSTL~\cite{Huang_2021_ICCV}, 3DLocal~\cite{Huang_2021_ICCV2}, and GaitGL~\cite{Lin_2021_ICCV}. All the results are averaged on the 11 gallery views and the identical views are excluded. As shown in~\cref{mainresults}, the accuracies of CL and the Mean condition (i.e., the average performance of NM, BG, CL) are remarkably improved by at least 3.8\% and 1.4\%, respectively. And the performance on NM and BG conditions is comparable with the state-of-the-art methods. Furthermore, as for the cross-view cases, GaitMPL achieves almost comprehensively better performance on all viewpoints with the small resolution of $64 \times 44$. This huge performance improvement demonstrates the necessity of tackling the two types of hard samples in gait recognition. Besides, it also indicates that the easy-to-hard progressive learning is mainly beneficial to the performance on the most challenging condition (CL).

\begin{table}[t]
  \caption{The averaged rank-1 accuracies on OU-MVLP. Identical views are excluded. GaitMPL adopts MT3D and GaitGL as the feature extractor of GaitMPL(M) and GaitMPL(G), respectively.}
  \scalebox{0.85}[0.8]{
  \scriptsize
  
  \begin{tabular}{c|ccccccc}
    \toprule
  \multicolumn{1}{c|}{\multirow{2}{*}{Probe}} & \multicolumn{7}{c}{Gallery All 14 Views}                     \\ \cmidrule {2-8} 
  \multicolumn{1}{c|}{}                       & GaitSet & GaitPart & GLN & MT3D & GaitGL & \textbf{GaitMPL(M)} & \textbf{GaitMPL(G)} \\ \midrule
  0°                                           &   79.5   &   82.6&  83.8     &  82.8   &   84.9     &   \textbf{83.9}      & \textbf{84.7}        \\
  15°                                          & 87.9        & 88.9         & 90.0    &   89.1   &  90.2      & \textbf{90.1}        &  \textbf{91.3}       \\
  30°                                          &  89.9       &   90.8       & 91.0    &  91.0    & 91.1       &   \textbf{91.3}       &  \textbf{92.2}       \\
  45°                                          &90.2         & 91.0        &  91.2   &  91.3    &  91.5      &  \textbf{91.5}        &  \textbf{92.4}       \\
  60°                                          &         88.1 &89.7 &90.3 &90.8 &   91.1     &  \textbf{91.2}        &  \textbf{91.9}       \\
  75°                                          &         88.7 & 89.9 & 90.0 & 89.3 & 90.8       &  \textbf{90.6}        & \textbf{91.3}         \\
  90°                                          &         87.8 & 89.5 & 89.4 & 89.3 & 90.3     &  \textbf{90.1}        & \textbf{91.5}        \\
  180°                                         &        81.7 &85.2 &85.3 &85.5 &   88.5     &    \textbf{85.3}      &   \textbf{86.1}      \\
  195°                                         &        86.7 &88.1 & 89.1 &88.2 & 88.6       &  \textbf{89.3}        &    \textbf{90.5}     \\
  210°                                         &         89.0 &90.0 & 90.5 &89.8 & 90.3       &  \textbf{90.7}        &  \textbf{91.3}       \\
  225°                                         &         89.3 &90.1 &90.6 &90.2 &  90.4     &   \textbf{90.7}       &   \textbf{91.6}      \\
  240°                                         &         87.2 &89.0 &89.6 & 88.7 & 89.6      &    \textbf{90.7}      &   \textbf{91.5}      \\
  255°                                         &           87.8 &89.1 &89.3 &89.3 &  89.5     &    \textbf{89.8}      &   \textbf{91.1}      \\
  270°                                         &          86.2 &88.2 & 88.5 &88.4 &   88.8     &    \textbf{88.9}      &    \textbf{90.0}     \\ \hline
  Mean                                        &          87.1 &88.7 & 89.2 &88.8 &   89.7     &    \textbf{89.6}      &  \textbf{90.6}       \\ \bottomrule
  \end{tabular}}
  \label{oumvlp_res}
  \end{table}

    \begin{table}[t]
        \scriptsize
        \centering
        \caption{The performance of methods integrated with DRPL and GSAM on CASIA-B, D and G denote DRPL and GSAM, respectively. \\``++'' denotes the methods that are equipped with DRPL and GSAM simultaneously.}
        \scalebox{1}[0.85]{
        \begin{tabular}{l|cccc}
        \toprule
        Method           & NM   & BG   & CL   & Mean  \\ \hline
        GaitSet          & 94.2 & 88.2 & 71.8 & 84.7 \\
        GaitSet + D      & 95.3          & 88.7          & 73.7          & 85.9          \\
        GaitSet + G      & 94.6          & 89.9          & 74.6          & 86.4          \\
        \textbf{GaitSet++}  & \textbf{95.6} & \textbf{88.7} & \textbf{80.9} & \textbf{88.4} \\ \hline
        GaitPart         & 96.1 & 92.1 & 78.2 & 88.8 \\
        GaitPart + D     & 95.5          & 91.3          & 82.8          & 89.9          \\
        GaitPart + G     & 96.8          & 92.4          & 80.0          & 89.7          \\
        \textbf{GaitPart++} & \textbf{95.7} & \textbf{91.9} & \textbf{83.8} & \textbf{90.5} \\ \hline
        MT3D             & 96.6 & 93.0 & 82.2 & 90.6 \\
        MT3D + D         & 95.1          & 92.5          & 86.8          & 91.5          \\
        MT3D + G         & 97.2          & 92.9          & 83.9          & 91.3          \\
        \textbf{MT3D++}     & \textbf{95.5} & \textbf{92.9} & \textbf{87.9} & \textbf{92.1}\\ \hline
        GLN        & 96.9 & 94.0 & 77.5 & 89.5 \\
        GLN + D     & 96.5 & 93.4 & 83.1 & 91.0 \\
        GLN + G      & 97.3 & 94.5 & 79.3 & 90.4 \\
        \textbf{GLN++} & \textbf{97.1} & \textbf{93.8} & \textbf{84.5} & \textbf{91.8} \\ \hline
        GaitGL         & 97.4 & 94.5 & 83.6 & 91.8 \\
        GaitGL + D        & 97.2 & 94.1 & 87.5 & 92.9 \\
        GaitGL + G        & 97.8 & 95.0 & 84.8 & 92.5 \\
        \textbf{GaitGL++} & \textbf{97.5} & \textbf{94.5} & \textbf{88.0} & \textbf{93.3} \\ 
    
        \bottomrule
        \end{tabular}}
        \label{ab_kecheng}
        \end{table}
    
\subsubsection{OU-MVLP} 
To verify the generalizability of GaitMPL under the large-scale scenario, we also conduct the experiments on the largest gait dataset, i.e., OU-MVLP. As shown in~\cref{oumvlp_res}, GaitMPL outperforms the state-of-the-art methods (e.g., GEINet~\cite{thapar2018vgr}, GaitSet, GaitPart, MT3D) by a considerable margin on wide viewpoints, which demonstrates the effectiveness of GaitMPL under the large-scale data scenario.

\subsection{Ablation Experiments}
In this part, firstly, we conduct experiments on several different kinds of methods to validate the effectiveness and generalizability of DRPL and GSAM. Secondly, we further analyze the effects and the hyperparameters of proposed modules based on GaitSet for its flexibility and simplicity.

\subsubsection{The Effectiveness and Generalizability of DRPL and GSAM}
To verify the effectiveness and generalizability of DRPL and GSAM, we integrate them into several different kinds of state-of-the-art methods and test their performance on CASIA-B and OU-MVLP, respectively.

% \usepackage{multirow}

\begin{table}[t]
    \scriptsize
    \centering
    \caption{The performance of methods integrated with DRPL and GSAM on OU-MVLP, which are marked with ``++''.}
    \scalebox{0.7}[0.8]{
    \begin{tabular}{l|cc|cc|cc|cc}
    \toprule
    \multicolumn{1}{l|}{\multirow{2}{*}{Probe}} & \multicolumn{8}{c}{Gallery All 14 Views}                    \\ \cmidrule{2-9} 
    \multicolumn{1}{l|}{}                       & GaitSet & \textbf{GaitSet++} & GaitPart & \textbf{GaitPart++} & MT3D & \textbf{MT3D++}  &GLN &\textbf{GLN++}\\ \hline
    Mean                                        & 87.3    &\textbf{88.2}      & 88.7     &\textbf{89.1}       & 88.8 & \textbf{89.6} & 89.2&\textbf{90.3}  \\
    \bottomrule
    \end{tabular}}
    \label{ou_abc}
    \end{table}

As shown in~\cref{ab_kecheng}, with the help of DRPL, the performance of state-of-the-art methods on CASIA-B is boosted considerably, especially on the most challenging condition (CL) and the average condition. 

As we have stated in~\cref{DRCL}, there is growing attention on the hard samples during the training process. And the performance gain indicates that the hard sample issue in gait recognition is mainly derived from the clothing condition. In other words, DRPL relatively pays more attention to the CL condition compared with other conditions (e.g., NM and BG) for better average accuracy. Therefore, the performance on NM and BG may drop slightly for a trade-off.
Furthermore, under the guidance of GSAM, the performance of the methods is improved by a considerable margin for all three conditions. 

A similar conclusion can be drawn from~\cref{ou_abc} on the largest dataset OU-MVLP, where state-of-the-art methods equipped with DRPL and GSAM (marked with ``++'') also achieve considerable performance gain on large-scale data scenarios.

These results above show that GSAM and DRPL could improve the performance by a considerable margin. More importantly, both DRPL and GSAM are independent of any network architecture, and they do not introduce extra cost at the inference phase. They are of considerable effectiveness and generalizability on mainstream state-of-the-art methods.

\subsubsection{The Analysis of the Components in DRPL}
\begin{table}[t]
    \centering
    \scriptsize
    \caption{Performance of different implementations for attention weight calculation in DRPL.}
    \scalebox{1}[0.8]{
    \begin{tabular}{l|cccc}
    \toprule
    Strategy            & \multicolumn{1}{c}{NM} & \multicolumn{1}{c}{BG} & \multicolumn{1}{c}{CL}     & \multicolumn{1}{c}{Mean}    \\ \midrule
    Baseline    & 94.2                 & 88.2                  & 71.8                      & \multicolumn{1}{c}{84.7} \\
    Thresholding      & 94.4                 & 87.6                  & 73.5                        & 85.1                      \\ 
    Dynamic Polynomial  & 94.4                    & 88.1                  & \textbf{74.8}                      & 85.7                      \\ 
    Exponential         & 94.6                  & 87.8                  & \multicolumn{1}{l}{72.8} & 85.1                      \\ 
    \textbf{Ours} & \textbf{95.3}                  & \textbf{88.8}                  & 73.7                      & \multicolumn{1}{c}{\textbf{85.9}}  \\ \bottomrule
    \end{tabular}}
    \label{ab_reweighting}
    \end{table}

\begin{table}[t]
    \centering
    \scriptsize
    \caption{Performance under different objective functions in DRPL.\\ SF denotes the smoothness factor.}
    \scalebox{0.9}[0.8]{
    \begin{tabular}{c|l|cccc}
    \toprule
    Index&Strategy                      & NM   & BG   & CL   & Mean  \\
    \midrule
    \ding{172}&Vanilla BH                    & 17.1 & 12.1 & 7.8  & 12.4 \\
    \ding{173}&DRPL-BH                       & 91.1 & 83.6 & 67.5 & 81.1 \\
    \ding{174}&Vanilla BH + BA (w/o SF)              & 93.5 & 86.5 & 71.9 & 83.9 \\
    \ding{175}&DRPL-BH + BA (w/o SF)                 & 94.8 & 87.4 & 72.4 & 84.9 \\
    \ding{176}&Vanilla BH + BA (w/ SF)&     93.6  & 86.5 & 72.1 & 84.0  \\
    \ding{177}&\textbf{DRPL-BH + BA (w/ SF)}  & \textbf{95.9} & \textbf{88.8} & \textbf{73.7} & \textbf{85.9}\\
    \bottomrule
    \end{tabular}}
    \label{smooth_DRCL} 
    \end{table}
\begin{table}[t]
    \centering
    \scriptsize
    \caption{The performance with different memory elements in GSAM.}
    \scalebox{1}[0.8]{
    \begin{tabular}{l|ccccc}
    \toprule
    Element     & NM   & BG   & CL   & Mean  & COST\\ \midrule
    Baseline     & 94.2 & 88.2 & 71.8 & 87.5 & 1$\times$ \\
    Hardest & 94.3 & 88.4 & 73.3 & 85.3 & 1$\times$\\
    Random     & 94.2 & 87.5 & 72.1 & 84.6 & 1$\times$\\
    Multi-Seqs & 94.4 & 89.3 & 73.2 & 85.6 &8$\times$ \\
    \textbf{Ours}       &\textbf{94.6} & \textbf{89.9} & \textbf{74.6} & \textbf{86.4} & 1$\times$\\
    \bottomrule
    \end{tabular}}
    \label{ab_content}
    \end{table}

\paragraph{The Calculation of Attention Weight in DRPL} Except for the dynamic exponential calculation for hardness metric adopted in~\cref{metric}, the calculation of weight distribution has plenty of implementations with different strategies~\cite{soviany2021curriculum}. To explore the effects of the calculation approaches for gait recognition, we compare our proposed dynamic exponential calculation with the following several commonly used approaches: \romannumeral1) Thresholding: filter out the samples under the specific threshold and associate the rest samples with normalized attention weights by Softmax; \romannumeral2) Dynamic Polynomial: calculate the weights in a polynomial fashion like $\sigma(d(F(x_i),F(x))^{\delta_t})$; \romannumeral3) Exponential: a static case that $\delta_t \equiv 1$ in~\cref{metric}, e.g., the positive weights are calculated by $\sigma(\exp{(d(F(x_i),F(x)))})$. As shown in~\cref{ab_reweighting}, all these alternative approaches improve the performance effectively compared with the baseline model, showing the effectiveness of the easy-to-hard progressive learning paradigm. Besides, the dynamic exponential calculation strategy obtains the best performance, which is applied in our DRPL module.

\paragraph{The Selection of Objective Function in DRPL} Hard sample mining through batch-hard-like loss inevitably introduces instability in the training process. The smoothness factor is proposed to stabilize the training using DRPL-BH and BA triplet loss. To show the effectiveness and stability of loss function described in~\cref{loss_drpl}, we conduct experiments with the combination of vanilla batch-hard triplet loss, batch-all triplet loss, $\mathcal{L}_{DRPL-BH}$ and, the smoothness factor.

The results in~\cref{smooth_DRCL} indicate that: \romannumeral1) \ding{172}\ding{173}, \ding{174}\ding{175} and \ding{176}\ding{177} indicate  DRPL-BH is more stable than vanilla BH, which implies that DRPL-BH is easier to smooth. \romannumeral2) \ding{172}\ding{174} and \ding{173}\ding{175} shows BA loss could stabilize the training process, but equal weight fusion only improves 0.2\% compared to BA. Thus, it is necessary to progressively increase the influence of unstable hard sample mining. \romannumeral3) \ding{174}\ding{176} and \ding{175}\ding{177} imply that the smoothness factor is compatible with DRPL-BH and hardly enhances vanilla BH, which needs a more complex smoothness operation. The reason may lie in that vanilla BH only selects the easiest and hardest sample pair, which would confuse the training due to the randomness and instability. DRPL-BH integrates all reweighted samples’ information, which could be stabilized by easier operation (i.e., a simple function).

\subsubsection{The Analysis of the Components in GSAM}

\paragraph{The Implementation of Memory Elements in GSAM} To demonstrate the effectiveness and compactness of the proposed centroid memory elements, we conduct several experiments under different cases: \romannumeral1) GaitSet without memory mechanism (Baseline); \romannumeral2) using the hardest sample of each ID as memory elements (termed as Hardest); \romannumeral3) using a random sample of each ID as the memory elements (termed as Random); \romannumeral4) using 8 random samples of each ID as the memory elements (termed as Multi-Seqs); \romannumeral5) our proposed centroid representation. As shown in~\cref{ab_content}, memory elements using the proposed centroid representation are more efficient. It could maintain only one feature for each ID while obtaining better performance than other methods that maintain multiple features.

\begin{table}[t]
    \centering
    \scriptsize
    \caption{The performance of different update strategies for memory bank.}
    \scalebox{1}[0.8]{
    \begin{tabular}{c|cccc}
    \toprule
    Method & NM & BG & CL & Mean \\
    \midrule
    Baseline & 94.2   & 88.2   &71.8    & 84.7    \\
    No update & 93.1   & 87.8   & 70.7    & 83.9 \\
    Hard update & 94.1   &  88.7  & 72.5   &  85.1  \\
    Momentum update (PU) &  94.4          & 88.2          & 73.8          & 85.5    \\
    \textbf{GSAM (PU+AU)}   &  \textbf{94.6}  & \textbf{89.9}   &  \textbf{74.6}  & \textbf{86.4}    \\
        \bottomrule
    \end{tabular}}
    \label{ab_sp}
    \end{table}
\paragraph{Different Update Strategy of GSAM} As shown in~\cref{ab_sp}, the results indicate that \romannumeral1) If not updating the memory, the model would be trained with outdated features, resulting in performance degradation. \romannumeral2) As MoCo~\cite{he2020momentum} points out that representations' consistency is very important, the momentum update strategy is better than hard update (directly replace).  \romannumeral3) The update strategy of GSAM (PU+AU) outperforms the vanilla momentum update strategy (PU) alone, which indicates structure-aligned strategy (AU) can effectively reduce the feature shift caused by asynchronous updates of the features in the memory bank.

\begin{figure}[t]
    \begin{center}
       \includegraphics[width=0.25\textwidth]{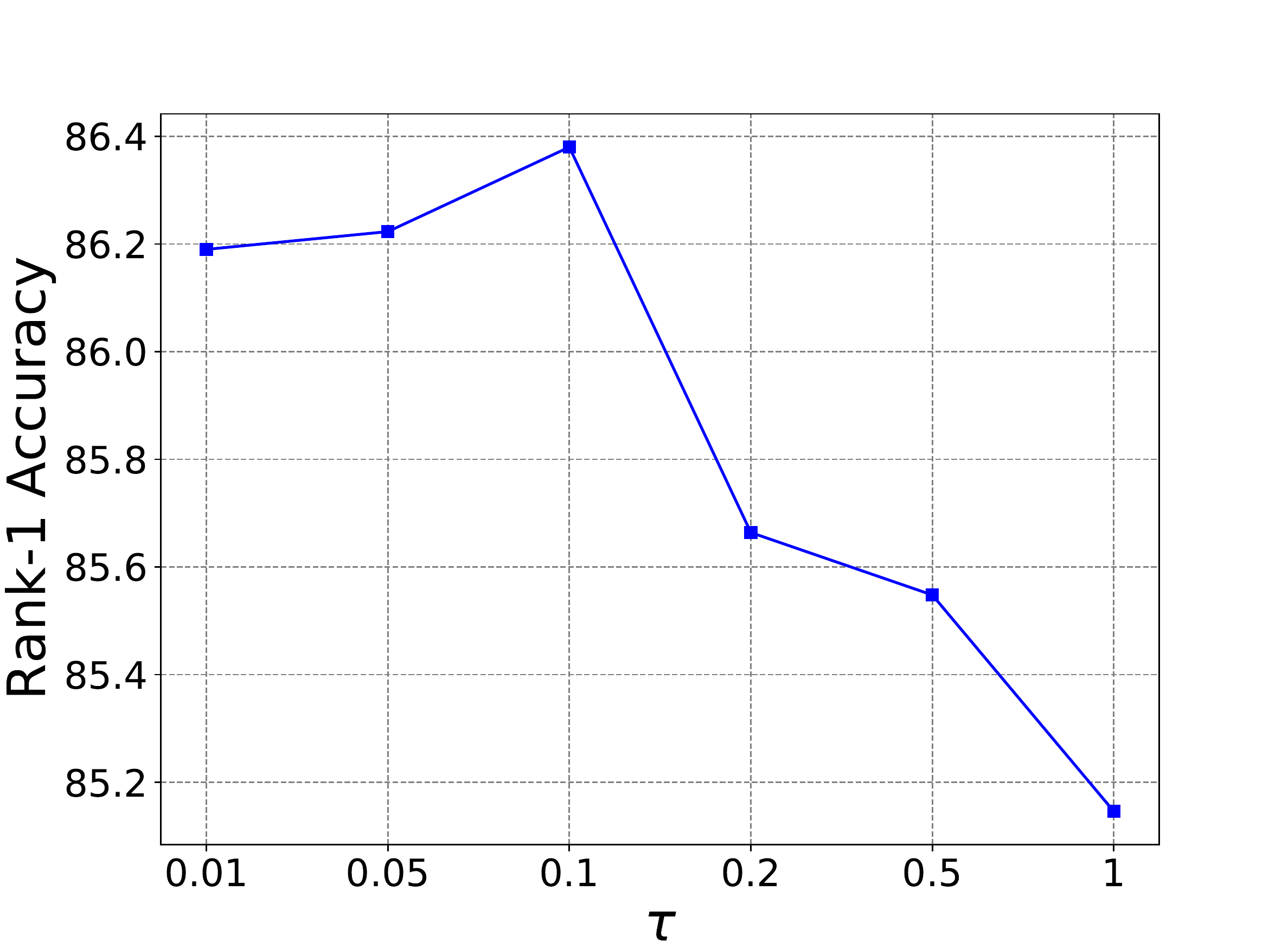}
    \end{center}
    \caption{The effect of temperature $\tau$ in GSAM. The best performance is achieved when $\tau=0.1$.}
    \label{fig:temperature}
 \end{figure}
\paragraph{The Selection of Temperature $\tau$} The temperature $\tau$ in GSAM is an important hyperparameter to balance the feature distribution. Though the features are normalized, several works~\cite{li2020joint, 9018132, 9336268, wu2020tracklet} show the significance of this hyperparameter. And~\cref{fig:temperature} shows the effects of temperature $\tau$ in~\cref{lmemory} and it achieves the best performance when $\tau=0.1$. On the whole, small $\tau$ makes the probability distribution closer to one-hot distribution and enlarges the discrepancy between items, which is more discriminative and obtains better performance. However, if $\tau$ is too small, it makes the training process unstable to the noise and damages the performance due to the numerical stability.

\begin{table}[t]
\centering
\scriptsize
\caption{Performance of different fusion factors \\
between DRPL and GSAM.}
\scalebox{1}[0.8]{
\begin{tabular}{l|llll}
\toprule
Strategy      & NM   & BG            & CL            & Mean           \\
\midrule
Baseline     & 94.2 & 88.2 & 71.8 & 84.7 \\
Equal weights & 95.1 & 88.4          & 79.2         & 87.6          \\
Multi Steps   & 95.2 & 88.3          & 80.1          & 87.9          \\
Logarithmic            & 95.5 & 88.5          & 80.5          & 88.2          \\
Exponential           & 95.5 & 88.1          & 80.6          & 88.1          \\
Polynomial         & 95.3 & 88.6          & 79.5          & 87.8         \\
\textbf{Sine}           & \textbf{95.6} & \textbf{88.7} & \textbf{80.9} & \textbf{88.4} \\
\bottomrule
\end{tabular}}
\label{ab_fusion}
\end{table}

\subsubsection{The Fusion Factor of DRPL and GSAM}
Following the design principle described in~\cref{training}, we adopt several different kinds of common functions on $\gamma_t$. And the results are shown in~\cref{ab_fusion}, the sine function is more suitable for gait recognition and outperforms the other forms of fusion factor by a huge margin. Therefore, we adopt the sine function as the fusion factor in our framework.

\label{ablation_exp}

\section{Conclusion}
In this paper, we propose a novel progressive learning network named Memory-augmented Progressive Learning (GaitMPL) to tackle the hard sample issue in gait recognition. GaitMPL is composed of two modules, i.e., Dynamic Reweighting Progressive Learning (DRPL) and Global Structure-Aligned Memory bank (GSAM). DRPL aims at alleviating the inconsistency of the learning objective between the easy samples and the hard samples by easy-to-hard progressive learning. And GSAM introduces the memory mechanism to augment the progressive learning process and uses a structure-aligned strategy to maintain the reliability of features in the memory bank.
Experiments on two mainstream benchmarks (CASIA-B and OU-MVLP) indicate the effectiveness and generalizability of GaitMPL and its components. Besides, the two components of GaitMPL can be integrated into most of the existing methods to achieve promising performance gain while introducing no extra costs in the inference phase.

\section*{Acknowledgements}
This work is supported in part by Zhejiang Provincial Natural Science Foundation of China under Grant LR19F020004, National Key Research and Development Program of China under Grant 2020AAA0107400, National Natural Science Foundation of China under Grant U20A20222, and Zhejiang University K.P.Chao's High Technology Development Foundation.

\IEEEpeerreviewmaketitle

\ifCLASSOPTIONcaptionsoff
  \newpage
\fi

\bibliographystyle{IEEEtran}
\bibliography{main}

\end{document}